\documentclass[]{sjtu_sai}
\usepackage[toc,page,header]{appendix}
\usepackage{minitoc}
\usepackage{algorithm}
\usepackage{algpseudocode}
\usepackage{booktabs}
\usepackage{multirow}
\usepackage{siunitx}
\usepackage{makecell}
\usepackage{mathtools}
\usepackage{tabularx,array}
\usepackage[table]{xcolor}
\usepackage{graphicx}
\usepackage{wrapfig}
\usepackage{xspace}
\usepackage{amsmath}
\usepackage{amsfonts}
\usepackage{hyperref}
\usepackage{url}
\usepackage{xspace}
\usepackage{subcaption}
\usepackage{caption}
\usepackage{adjustbox}
\usepackage{placeins}
\usepackage{eso-pic}
\usepackage{tikz}
\newcommand{\github}{\raisebox{-1.5pt}{\includegraphics[height=1.05em]{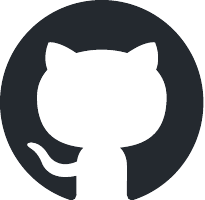}}\xspace}

\newcommand{\homepage}{\raisebox{-1.5pt}{\includegraphics[height=1.05em]{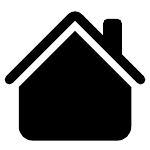}}\xspace}

\newcommand{\headerlogo}[2][15mm]{%
  \raisebox{-0.5\height}{%
    \includegraphics[
      width=24mm,
      height=#1,
      keepaspectratio
    ]{#2}%
  }%
}

\newcommand{\method}{WAVE-0\xspace}
\makeatletter
\newcount\sai@affilcount
\renewcommand\affiliation[2][]{%
  \begingroup
  \def\sep{}%
  \ifnum\sai@affilcount=0\relax
    \def\sep{}%
  \else\ifnum\sai@affilcount=1\relax
    \def\sep{\par}% break after the first affiliation
  \else
    \def\sep{\quad}% spacing within the second line
  \fi\fi
  \global\advance\sai@affilcount by 1\relax
  \let\protect\@unexpandable@protect
  \xdef\affiliationlist{\expandafter{\affiliationlist}\sep \affiliationformat[#1]{#2}}%
  \endgroup
}
\makeatother

\newcommand{\mytoprule}{
    \toprule
    \noalign{\vspace{-0.2mm}}
}
\newcommand{\mymidrule}{
    \noalign{\vspace{-0.8mm}}
    \midrule
    \noalign{\vspace{-1mm}}
}
\newcommand{\mybottomrule}{
    \noalign{\vspace{-0.6mm}}
    \bottomrule
}
\definecolor{oursgray}{gray}{0.95}
\definecolor{oursgreen}{rgb}{0.90, 0.96, 0.90}
\definecolor{oursblue}{rgb}{0.918, 0.941, 0.988}
\definecolor{improvementgreen}{rgb}{0.0, 0.4, 0.15}
\newcommand{\impv}[1]{\textcolor{improvementgreen}{\textsl{#1}}}
\definecolor{decreasered}{rgb}{0.65, 0.15, 0.15}
\newcommand{\dropv}[1]{\textcolor{decreasered}{\textsl{#1}}}

\title{WAVE-0: Bridging VLA Policies and WAMs\\ with Lightweight Latent Future Prediction}

\author[1]{Haomin Wang}
\author[1,3]{Anqiong Ji}
\author[2,4]{Ke Wang}
\author[3]{Zhongshan Liu}
\author[3]{Ru Ying}
\author[3]{Ran Zheng}
\author[1\dagger]{Zhipeng Zhang}

\affiliation[1]{AutoLab, School of Artificial Intelligence, Shanghai Jiao Tong University}
\affiliation[2]{Awaken Robotics}
\affiliation[3]{Baidu AI Cloud}
\affiliation[4]{Zhejiang University}
\contribution[\dagger]{Corresponding Author}

\abstract{Vision–language–action (VLA) policies excel at generalist robotic manipulation but generate actions without explicitly modeling their environmental consequences. World action models (WAMs) incorporate future prediction, yet most rely on costly video generation, while latent alternatives either use predicted futures only as auxiliary targets or infer latent actions from suboptimal visual transitions through costly large-scale video pretraining. We introduce \textbf{\method} (\textbf{W}orld-model-\textbf{A}ugmented \textbf{V}LA \textbf{E}xecution), a framework that bridges VLAs and WAMs through a lightweight latent world model. By connecting the semantic understanding of pretrained VLAs with explicit future prediction, \method allows anticipated action consequences to directly guide action generation. Built on a pretrained VLA, its 257M world model translates VLM-generated latent action plans into predicted future representations that condition the action expert, unifying latent planning, future prediction, and action generation. The world model is trained directly on ground-truth robot actions with a counterfactual ranking objective, requiring neither pixel reconstruction nor large-scale video pretraining. \method achieves success rates of 91.47\% on RoboTwin 2.0 and 98.7\% on LIBERO, along with 88.0\%, 61.4\%, and 75.0\% on LIBERO-Plus, LIBERO-Pro, and LIBERO-Para, respectively. On real-world tasks, it improves the success rate of $\pi_{0.5}$ from 38.3\% to 73.3\%. These results demonstrate strong manipulation performance and generalization across diverse distribution shifts.}

\checkdata[\homepage Project Page]{\url{https://hmwang2002.github.io/release/wave0/}}
\checkdata[\github GitHub Repo]{\url{https://github.com/AutoLab-SAI-SJTU/WAVE-0}}
\begin{document}
% 星号表示仅添加到下一张输出页面，即首页
\AddToShipoutPictureFG*{%
  \begin{tikzpicture}[remember picture, overlay]
    \node[
      anchor=north west,
      inner sep=0pt,
      xshift=25mm,
      yshift=-12mm
    ] at (current page.north west) {%
      \headerlogo[15mm]{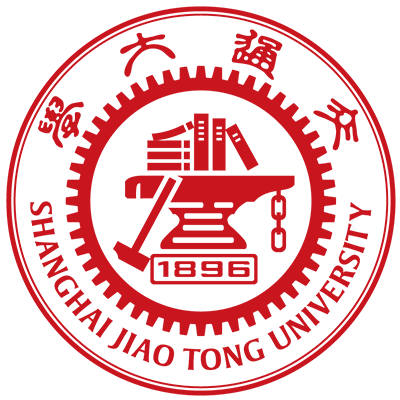}%
      \hspace{1mm}%
      \headerlogo[17mm]{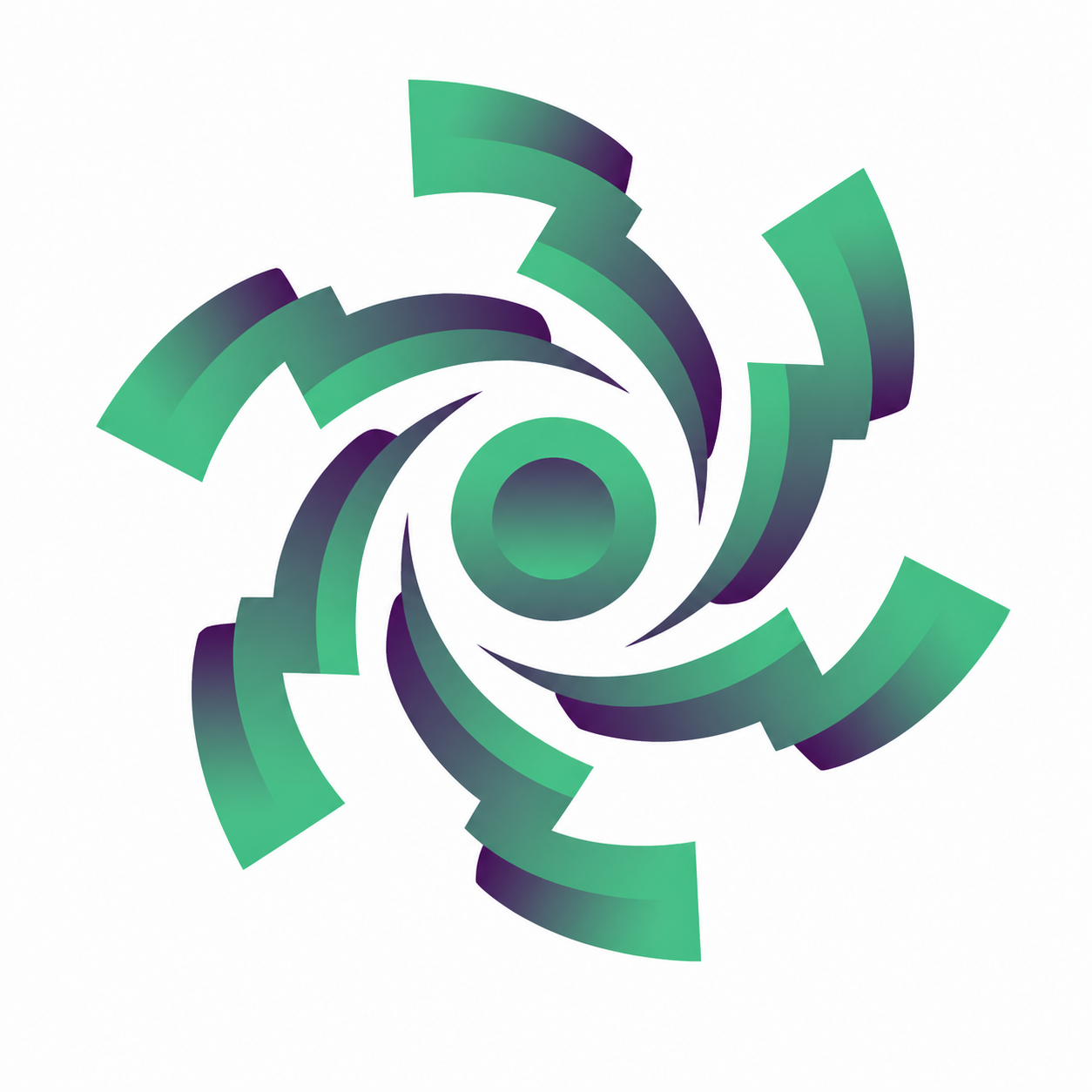}%
      \hspace{1mm}%
      \headerlogo[11mm]{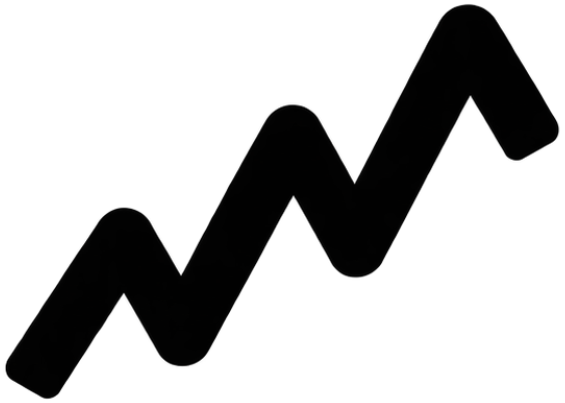}%
      \hspace{1mm}%
      \headerlogo[14mm]{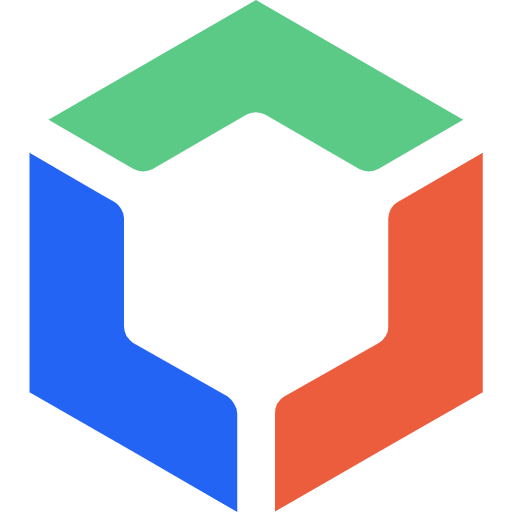}%
      \hspace{1mm}%
      \headerlogo[16mm]{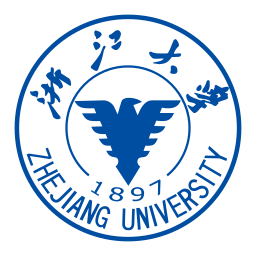}%
    };
  \end{tikzpicture}%
}

% 后面保留原来的标题生成代码
\maketitle

\section{Introduction}
\label{sec:intro}

Vision–language–action (VLA) policies have made substantial progress in general robotic manipulation by combining the semantic knowledge of pretrained vision–language models with action supervision from robot demonstrations~\citep{kim2024openvla,pi0,pi05,bjorck2025gr00t,brohan2022rt,brohan2023rt,liu2025rdt}. While effective, this direct observation-to-action mapping does not explicitly capture how actions transform the environment and thereby shape subsequent decisions. World action models (WAMs) address this limitation by incorporating environmental dynamics into action generation, enabling policies to anticipate the future consequences of their actions~\citep{ye2026world,bi2026motus,li2026causal}. Most existing WAMs build on large-scale video generation models, whose pixel-level prediction spends substantial computation on visual details irrelevant to action selection and limits inference efficiency. Recent efforts have tried to reduce this overhead~\citep{yuan2026fast,cai2026aha}, but they still rely on video generation backbones that devote considerable capacity to visual reconstruction.

A promising alternative is to model environmental dynamics in a compact latent space, allowing future prediction to focus on information relevant to action selection rather than visual reconstruction. VLA-JEPA~\citep{sun2026vla} learns latent actions through future-state prediction, while JEPA-WAM~\citep{lin2026jepa} uses a shared predictor to model state transitions and generate actions. However, neither approach explicitly conditions action generation on predicted future states, leaving future prediction and action selection only loosely connected. LaWAM~\citep{chen2026lawam} guides action generation with predicted latent visual subgoals, but learns its latent actions indirectly from visual transitions through inverse dynamics. While this formulation enables world model pretraining on large-scale videos without action annotations, the inferred latent actions may summarize realized transitions rather than the executable actions that caused them. Consequently, a representation that explains an observed transition may be misaligned with the action a policy must generate before the transition occurs. Moreover, LaWAM pretrains its world model on approximately 4,500 hours of videos, followed by policy-integration pretraining and additional benchmark-specific training, resulting in substantial training costs. This raises a natural question: \textit{when domain-specific demonstrations already provide ground-truth executable actions, is such large-scale, indirect pretraining truly more effective than directly training the world model with those actions?}

\begin{figure}[t]
  \begin{center}
  \includegraphics[width=0.9\textwidth]{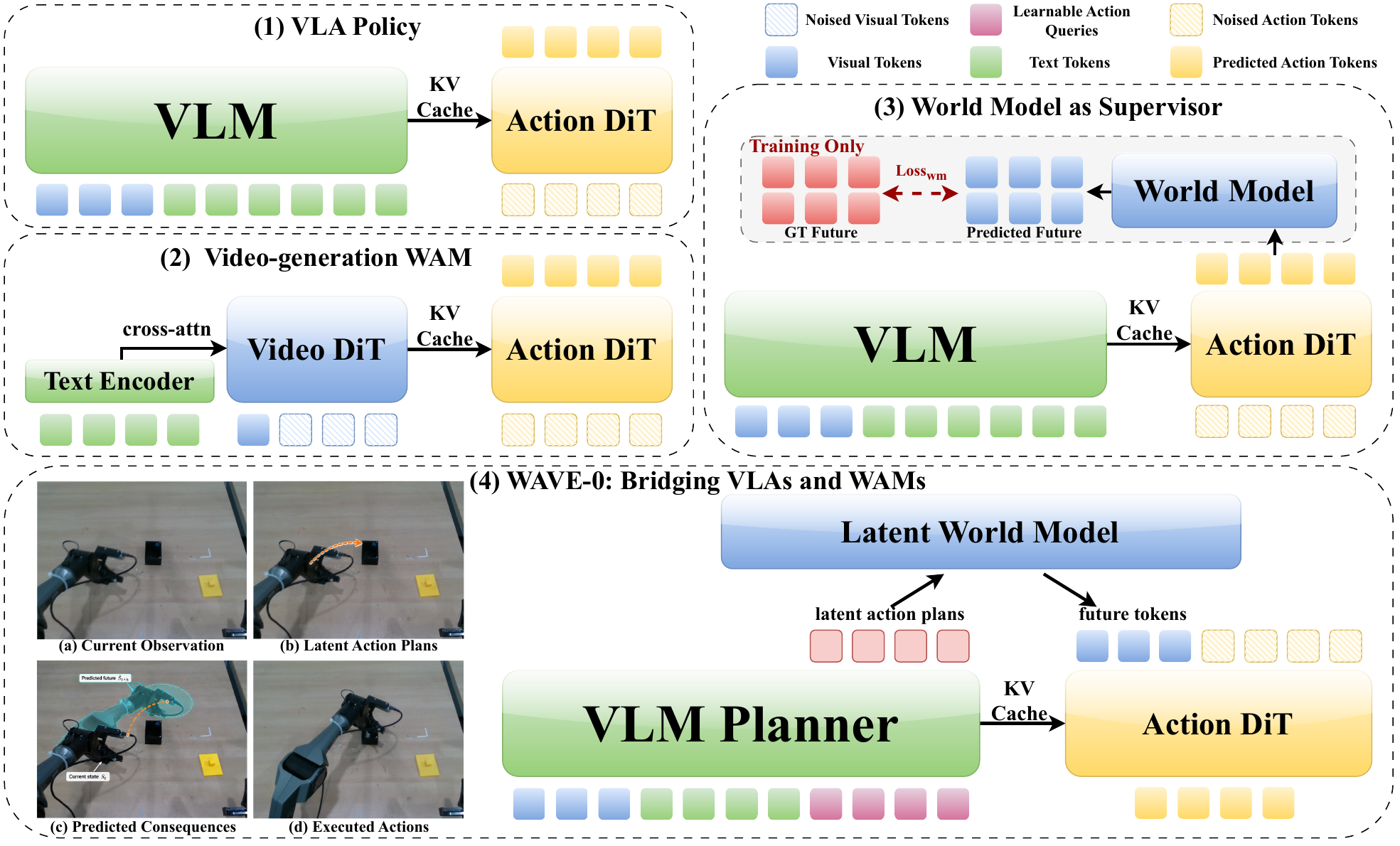}
  \caption{
\textbf{Comparison of representative robot policy paradigms.} (1) Standard VLAs map vision-language context directly to actions.
(2) Video-based WAMs model dynamics through costly video generation.
(3) A world model provides auxiliary supervision by predicting the consequences of generated actions during training, leaving the inference pathway unchanged.
(4) \method translates VLM-generated latent action plans into predicted future representations with a lightweight latent world model, conditioning the action expert during both training and inference.
  }
  \label{fig:comparison}
  \end{center}
  % \vspace{-1.5em}
\end{figure}

We answer this question by grounding future prediction directly in the ground-truth actions of robot trajectories. Our key insight is that demonstrations already pair each action sequence with its resulting observations, providing direct supervision on how actions change the environment. A world model trained with this supervision predicts futures tied to the executed actions rather than to latent actions inferred after the transition occurs, and requires no large-scale pretraining. Moreover, since a pretrained VLA already provides rich semantic understanding and a visual representation space, the world model only needs to predict how this representation evolves under a given action plan, which keeps it lightweight and avoids reconstructing future images. We therefore view semantic understanding in VLAs and future prediction in WAMs as complementary, where the former determines action intent and the latter anticipates its consequences. Building on this view, we introduce \textbf{\method} (\textbf{W}orld-model-\textbf{A}ugmented \textbf{V}LA \textbf{E}xecution), a framework that bridges VLAs and WAMs through action-grounded latent future prediction. As shown in Figure~\ref{fig:comparison}, our latent world model transforms VLM-generated latent action plans into predicted future visual representations that explicitly condition the action expert, yielding a unified pipeline of \emph{latent planning, future prediction, and action generation}.

\method follows a two-stage training strategy. In the first stage, we train the latent world model on action-labeled trajectories to predict future visual representations from current observations and ground-truth actions. Besides future and latent delta objectives, a counterfactual ranking objective encourages the model to distinguish the consequences of different action plans. In the second stage, we integrate the frozen world model into pretrained $\pi_{0.5}$~\citep{pi05} and align VLM-generated latent plans with the learned action representation space. The world model predicts future representations from these plans, and a lightweight resampler compresses them into future-conditioning tokens for the action expert. The planning branch guides action generation only through these predictions, while the expert retains access to the original VLM context. This design allows anticipated action consequences to explicitly guide action generation during both training and inference.

On RoboTwin 2.0~\citep{chen2025robotwin} and LIBERO~\citep{liu2023libero}, \method achieves average success rates of 91.47\% and 98.7\%, improving over $\pi_{0.5}$ by 11.09 and 1.8 percentage points, respectively. On the more challenging LIBERO-Plus~\citep{fei2025libero_plus} and LIBERO-Pro~\citep{zhou2025libero_pro}, which introduce variations in objects, layouts, camera viewpoints, and instructions, \method achieves 88.0\% and 61.4\%, outperforming all compared methods. On LIBERO-Para~\citep{kim2026libero_para}, it exceeds $\pi_{0.5}$ by 3.6 percentage points under instruction paraphrasing. These gains extend to three real-world tasks, where \method improves the average success rate of $\pi_{0.5}$ from 38.3\% to 73.3\%. Compared with LaWAM, \method is also more training-efficient, processing far fewer training examples across RoboTwin 2.0 and real-world tasks. Together, these results demonstrate that action-grounded latent future prediction not only improves a strong VLA with modest cost, but also substantially enhances its robustness to visual, spatial, and language distribution shifts.

Our contributions are threefold:

(1) We introduce \method, a framework that connects VLA semantic understanding with WAM future prediction through a 257M lightweight latent world model, enabling anticipated action consequences to guide action generation during training and inference.

(2) We develop a two-stage training strategy that learns action-dependent latent dynamics from ground-truth robot actions using counterfactual ranking, then integrates the frozen world model into a pretrained VLA through an explicit prediction bottleneck.

(3) Evaluations on RoboTwin 2.0, LIBERO, and its generalization benchmarks demonstrate \method's effectiveness and generalization under visual and language distribution shifts. Real-world experiments further demonstrate substantial improvements over $\pi_{0.5}$ across three manipulation tasks.
\section{Related Works}
\label{sec:related_works}

\subsection{Vision-Language-Action Policies}

Vision-language-action (VLA) policies have advanced generalist robotic manipulation by combining pretrained vision-language knowledge with robot action supervision~\citep{brohan2022rt,brohan2023rt,kim2024openvla,kim2025fine,team2025gemini,yang2026abot}. Diffusion-based and flow-based policies further support expressive continuous action generation across diverse tasks and embodiments~\citep{pi0,pi05,bjorck2025gr00t,liu2025rdt,wen2025dexvla}. Beyond action generation, embodied reasoning and explicit spatial representations have also been explored to enhance VLA policies~\citep{zawalski2024robotic,qu2025spatialvla}. Another line of work learns task-centric latent action representations from videos, enabling VLA pretraining across embodiments with reduced reliance on action annotations~\citep{ye2025latent,bu2025univla,yang2026learning}. Despite these advances, VLAs primarily map observations and instructions to actions, leaving action-dependent environmental dynamics implicit. This motivates integrating explicit future prediction into VLAs, allowing anticipated action consequences to directly guide action generation.

\subsection{Visual Generation-Based World Action Models}

World action models (WAMs) incorporate future-state prediction into action generation to explicitly model how robot actions affect the environment. Most existing WAMs build on pretrained video models that jointly predict future observations and actions~\citep{ye2026world,bi2026motus,kim2026cosmos,li2026causal,pai2025mimic,liang2025video,liao2026genie}. However, large backbones, modeling of action-irrelevant visual details, and iterative denoising incur substantial computational cost and latency. To reduce this cost, Fast-WAM~\citep{yuan2026fast} and GigaWorld-Policy~\citep{ye2026gigaworld} retain visual prediction only as training supervision and skip future generation at inference, while other methods reuse future information through asynchronous execution, sparse interactions, or single-pass caching~\citep{cai2026aha,zhao2026faster,zhang2026keep}. Beyond video generation, BagelVLA~\citep{hu2026bagelvla} builds on a unified model for understanding and generation to extract predictive visual features through single-step denoising, while ImageWAM~\citep{zhang2026imagewam} guides action generation using denoising features from an image-editing model without decoding future images at inference. However, visual generation backbones remain computationally demanding, motivating lightweight dynamics models that directly predict future visual representations without pixel-level reconstruction.

\subsection{Latent World Action Models}

Joint-embedding predictive architecture (JEPA) offers a framework for latent world action modeling by predicting representations without pixel reconstruction~\citep{lecun2022path,assran2023self,assran2025v,maes2026leworldmodel}. VLA-JEPA~\citep{sun2026vla} learns action-relevant latent representations from action-free human videos through world prediction, and then adapts them to robot control. JEPA-WAM~\citep{lin2026jepa} uses spatially structured targets that jointly encode current and future observations, providing dense transition supervision for policy learning. However, both methods use predictive supervision to shape policy representations without directly conditioning action generation on predicted visual representations. LaWAM~\citep{chen2026lawam} learns latent actions through visual inverse dynamics and uses them to predict visual subgoals that guide action generation. We posit that semantic understanding in pretrained VLAs and future prediction in WAMs are complementary capabilities that should jointly guide action generation. \method bridges these paradigms through a lightweight latent world model that learns to predict world state transitions from labeled robot action sequences. It translates VLM-generated latent action plans into future representations that explicitly guide the action expert during both training and inference.
\section{Method}
\label{sec:method}

\subsection{Problem Formulation}

We consider robot manipulation conditioned on a current observation and a task instruction. At time $t$, the robot receives an observation $o_t=(I_t,s_t)$, where $I_t=\{I_t^{(c)}\}_{c=1}^{C}$ contains images from $C$ cameras and $s_t$ is the proprioceptive state. Given an instruction $\ell$, the robot generates a continuous action chunk with an action horizon of $H$:
\begin{equation}
A_t=(a_t,\ldots,a_{t+H-1}), \qquad a_t\in\mathbb{R}^{d_a}.
\label{eq:action_chunk}
\end{equation}

We learn a policy from demonstrations to generate actions conditioned on the current observation and instruction:
\begin{equation}
A_t\sim\pi_\theta(\cdot\mid o_t,\ell).
\label{eq:action_policy}
\end{equation}

\method decomposes this process into latent action planning, future world state prediction, and future-conditioned action generation. Let $S_t=\mathcal E(I_t)$ denote the world state representation in a fixed visual embedding space. A VLM planner $f_\eta$ produces a latent action plan, a world model $g_\phi$ predicts its consequences, and an action expert $p_\psi$ generates actions using both the original context and the predicted future:
\begin{equation}
\widehat Z_a=f_\eta(o_t,\ell), \quad \widehat S_{t+h}=g_\phi(S_t,s_t,\widehat Z_a), \quad A_t\sim p_\psi(\cdot\mid o_t,\ell,\widehat S_{t+h}),
\label{eq:policy_decomposition}
\end{equation}
where $h$ denotes the future observation offset.

\subsection{Model Architecture}

\begin{figure}[htbp]
  \begin{center}
  \includegraphics[width=1.0\textwidth]{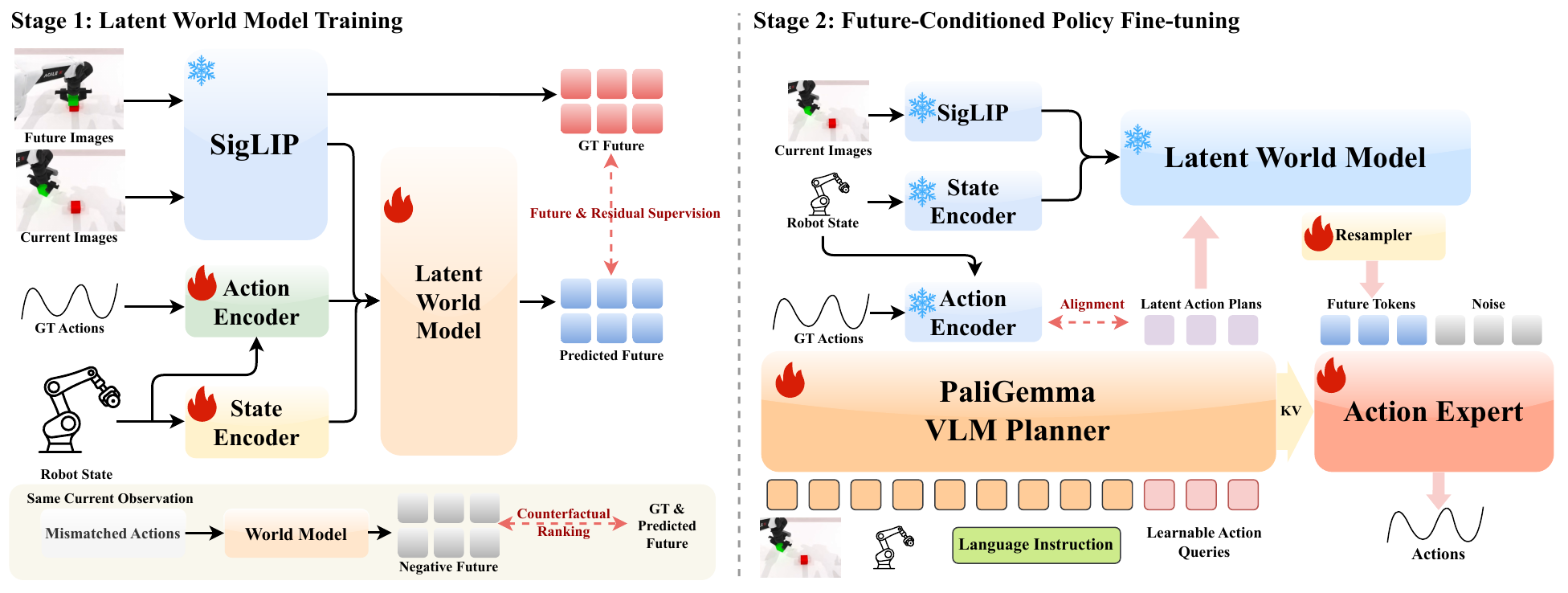}
  \caption{
\textbf{Overview of \method.} \textbf{Stage~1:} The latent world model learns action-conditioned dynamics in a fixed visual embedding space, trained with future-representation prediction, latent residual supervision, and counterfactual action ranking. \textbf{Stage~2:} The VLM generates latent action plans aligned with the representations from the frozen action encoder. The frozen world model predicts the visual consequences of these plans, and a resampler compresses them into future tokens that condition the action expert alongside the original VLM context.
  }
  \label{fig:method}
  \end{center}
  % \vspace{-2.5em}
\end{figure}

Built on pretrained $\pi_{0.5}$~\citep{pi05}, \method combines vision-language understanding with explicit dynamics prediction through latent action planning, future world state prediction, and future-conditioned action generation. The training procedure is conducted in two sequential stages. In the first stage, the latent world model jointly learns action representations and latent dynamics from visual observations, proprioceptive states, and ground-truth action sequences to predict future world states in a fixed visual embedding space. In the second stage, we integrate the frozen world model into $\pi_{0.5}$ and train the VLM to produce latent action plans aligned with the learned action representation space. The world model converts these plans into predicted future visual representations, which condition the action expert alongside the original vision-language context.

\textbf{Latent World Model.}~
We learn action-conditioned latent dynamics in a fixed visual embedding space. We obtain $S_t=\mathcal E(I_t)$ and $S_{t+h}=\mathcal E(I_{t+h})$ using the frozen SigLIP~\citep{zhai2023sigmoid} encoder and visual projector from pretrained $\pi_{0.5}$, followed by spatial pooling. An action encoder embeds the ground-truth action sequence conditioned on the current proprioceptive state, yielding $Z_a=\mathcal E_a(A_t,s_t)$, while a state encoder produces $z_s=\mathcal E_s(s_t)$. A Transformer with action-conditioned AdaLN-Zero blocks~\citep{peebles2023scalable} predicts a latent residual, which is added to the current representation to obtain the predicted future:
\begin{equation}
\Delta \widehat{S}=\mathcal W_\phi(S_t,z_s,Z_a),
\quad
\widehat S_{t+h}=S_t+\Delta \widehat{S}.
\label{eq:latent_world_prediction}
\end{equation}

We supervise the future representation with a cosine distance loss and the state change with a SmoothL1 loss:
\begin{equation}
\mathcal L_{\mathrm{future}}=d_{\cos}(\widehat S_{t+h},S_{t+h}), \quad \mathcal L_{\mathrm{delta}}=\operatorname{SmoothL1}(\Delta \widehat{S},S_{t+h}-S_t),
\label{eq:world_prediction_losses}
\end{equation}
where $d_{\cos}$ denotes mean token-wise cosine distance. 

To encourage action dependence, we introduce a counterfactual ranking loss. Keeping the current observation fixed, we replace the demonstrated actions with a mismatched sequence $\widetilde A_t$ to obtain:
\begin{equation}
\widehat S_{t+h}^{-}=S_t+\mathcal W_\phi(S_t,z_s,\mathcal E_a(\widetilde A_t,s_t)).
\label{eq:negative_world_prediction}
\end{equation}

The loss favors demonstrated actions over mismatched actions when predicting the same observed future:
\begin{equation}
\mathcal L_{\mathrm{rank}}=r\left[m+d_{\cos}(\widehat S_{t+h},S_{t+h})-d_{\cos}(\widehat S_{t+h}^{-},S_{t+h})\right]_+,
\label{eq:world_ranking_loss}
\end{equation}
where $[x]_+=\max(0,x)$, $m=0.1$, and $r\in\{0,1\}$ masks unchanged negative sequences.

We jointly optimize the action encoder, state encoder, and world model using the following objective:
\begin{equation}
\mathcal L_{\mathrm{stage1}}
=\mathcal L_{\mathrm{future}}
+0.5\,\mathcal L_{\mathrm{delta}}
+0.1\,\mathcal L_{\mathrm{rank}}.
\label{eq:stage1_objective}
\end{equation}

\textbf{Latent Action Planning.}~
We integrate the frozen latent world model into pretrained $\pi_{0.5}$ and train its PaliGemma VLM~\citep{beyer2024paligemma} to generate latent action plans in the action representation space learned in Stage 1. Specifically, we append learnable latent action queries $Q$ to the VLM input sequence and transform the output hidden states $U_Q$ into latent action tokens via $\widehat Z_a=\Pi_a(\operatorname{LN}(U_Q))$, where $\Pi_a$ is a learned linear projection. These tokens condition the frozen world model to predict future visual representations:
\begin{equation}
\widehat S_{t+h}
=S_t+\mathcal W_\phi(S_t,z_s,\widehat Z_a).
\label{eq:latent_action_planning}
\end{equation}

We supervise latent action planning by aligning $\widehat Z_a$ with the target representations $Z_a$ produced by the frozen action encoder:
\begin{equation}
\mathcal L_{\mathrm{align}}
=d_{\cos}(\widehat Z_a,Z_a)
+0.5\,\operatorname{SmoothL1}(\widehat Z_a,Z_a).
\label{eq:latent_action_alignment}
\end{equation}

To supervise the predicted visual consequences, we reuse the future representation and residual losses in Eq.~\ref{eq:world_prediction_losses}, evaluated with $\widehat Z_a$ as the action condition:
\begin{equation}
\mathcal L_{\mathrm{world}}=\mathcal L_{\mathrm{future}}+0.5\,\mathcal L_{\mathrm{delta}}.
\label{eq:world_consistency}
\end{equation}

\textbf{Future-Conditioned Action Generation.}~
A lightweight resampler pools and projects the predicted future representation into conditioning tokens $F_t=\mathcal R(\widehat S_{t+h})$. We prepend these tokens to the noisy action tokens in the action expert, enabling it to attend to both the original VLM context and the predicted future. We exclude latent-action query tokens from the context passed to the expert, so that the planning branch guides action generation only through its predicted visual consequences.

In Stage~2, we jointly optimize the VLM planner, future-conditioning resampler, and action expert while keeping the world model frozen:
\begin{equation}
\mathcal L_{\mathrm{stage2}}
=\mathcal L_{\mathrm{flow}}
+\lambda_{\mathrm{WAM}}
\left(
\mathcal L_{\mathrm{world}}
+\lambda_{\mathrm{align}}\mathcal L_{\mathrm{align}}
\right),
\label{eq:stage2_objective}
\end{equation}
where $\mathcal L_{\mathrm{flow}}$ is the flow-matching objective used in $\pi_{0.5}$, $\lambda_{\mathrm{WAM}}$ scales the auxiliary supervision, and $\lambda_{\mathrm{align}}$ controls the relative weight of latent-action alignment.

\section{Experiments}
\label{sec:exp}

\subsection{Implementation Details}

We initialize the VLM and action expert from pretrained $\pi_{0.5}$. For world modeling, we use a frozen copy of its SigLIP encoder and visual projector to extract 2048-dimensional visual features, which are spatially pooled into a $4\times4$ grid, yielding 16 tokens per camera. The latent world model processes these representations using a 12-layer Transformer with a hidden dimension of 1024, 16 attention heads, and an MLP dimension of 4096. The resampler then pools the predicted future representations to a $2\times2$ grid per camera and projects them into 12 conditioning tokens for the action expert, which retains the original flow-matching formulation of $\pi_{0.5}$.

The world model, action encoder, and state encoder together comprise approximately 257M trainable parameters in Stage~1, while the full \method contains approximately 4.3B parameters. We train with AdamW using a default peak learning rate of $10^{-4}$, weight decay of $0.01$, and a cosine learning rate schedule. Benchmark-specific training and inference configurations are provided in Appendix~\ref{appendix:training_evaluation}.

\begin{table}[htbp]
\centering
\caption{\textbf{Success rates (\%) on LIBERO and RoboTwin 2.0.} Bold and underlined values indicate the best and second-best results, respectively. $\Delta$ denotes improvement over $\pi_{0.5}$.}
\label{tab:libero_and_robotwin}
\begin{subtable}[t]{0.53\textwidth}
\centering
\caption{Results on LIBERO.}
\label{tab:libero}
\resizebox{\linewidth}{!}{%
\begin{tabular}{lccccc}
\mytoprule
\textbf{Method} & \textbf{Spatial} & \textbf{Object} & \textbf{Goal} & \textbf{Long} & \textbf{Avg.} \\
\mymidrule
OpenVLA & 84.7 & 88.4 & 79.2 & 53.7 & 76.5 \\
GR00T N1 & 94.4 & 97.6 & 93.0 & 90.6 & 93.9 \\
$\pi_0$ & 96.8 & 98.8 & 95.8 & 85.2 & 94.1 \\
Motus & 96.8 & \underline{99.8} & 96.6 & 97.6 & 97.7 \\
LingBot-VA & 98.5 & 99.6 & 97.2 & \textbf{98.5} & 98.5 \\
Fast-WAM & 98.2 & \textbf{100.0} & 97.0 & 95.2 & 97.6 \\
ImageWAM & 97.2 & 99.2 & \textbf{98.8} & \underline{98.4} & 98.4 \\
VLA-JEPA & 96.2 & 99.6 & 97.2 & 95.8 & 97.2 \\
JEPA-WAM & 95.6 & 99.4 & 97.2 & 94.6 & 96.7 \\
LaWAM & \textbf{99.4} & 99.6 & 98.4 & 97.0 & \underline{98.6} \\
\mymidrule
$\pi_{0.5}$ & \underline{98.8} & 98.2 & 98.0 & 92.4 & 96.9 \\
\rowcolor{oursgreen}
\textbf{WAVE-0 (Ours)} & 98.6 & \textbf{100.0} & \underline{98.6} & 97.6 & \textbf{98.7} \\
\textsl{$\Delta$ Improvement} & \dropv{-0.2} & \impv{+1.8} & \impv{+0.6} & \impv{+5.2} & \impv{+1.8} \\
\mybottomrule
\end{tabular}%
}
\end{subtable}
\hfill
\begin{subtable}[t]{0.45\textwidth}
\centering
\caption{Results on RoboTwin 2.0.}
\label{tab:robotwin}
\resizebox{\linewidth}{!}{%
\begin{tabular}{lccc}
\mytoprule
\textbf{Method} & \textbf{Clean} & \textbf{Randomized} & \textbf{Avg.} \\
\mymidrule
$\pi_0$ & 65.92 & 58.40 & 62.16 \\
ABot-M0 & 81.20 & 80.40 & 80.80 \\
Motus & 88.66 & 87.02 & 87.84 \\
GigaWorld-Policy & 86.36 & 85.04 & 85.70 \\
LingBot-VA & 92.90 & 91.50 & 92.20 \\
Fast-WAM & 91.88 & 91.78 & 91.83 \\
AHA-WAM & \textbf{93.40} & \underline{92.20} & \underline{92.80} \\
LaWAM & 92.64 & 89.80 & 91.22 \\
ImageWAM & \underline{93.20} & \textbf{93.56} & \textbf{93.38} \\
\mymidrule
$\pi_{0.5}$ & 81.12 & 79.64 & 80.38 \\
$\pi_{0.5}$ + WM & 89.48 & 89.54 & 89.51 \\
\rowcolor{oursgreen}
\textbf{WAVE-0 (Ours)} & 91.46 & 91.48 & 91.47 \\
\textsl{$\Delta$ Improvement} & \impv{+10.34} & \impv{+11.84} & \impv{+11.09} \\
\mybottomrule
\end{tabular}%
}
\end{subtable}
\end{table}

\subsection{Evaluation on LIBERO}
We evaluate on the four LIBERO suites: Spatial, Object, Goal, and Long, covering 40 tasks. Our training data comprise 1,693 trajectories and 273,465 frames across these suites. As shown in Table~\ref{tab:libero}, \method achieves an average success rate of 98.7\%, the highest among the compared methods, improving over $\pi_{0.5}$ by 1.8 percentage points. The largest gain occurs on the Long suite, where success increases from 92.4\% to 97.6\%. This suggests that anticipating action consequences is particularly beneficial for sequential manipulation, where the outcome of each action affects subsequent steps. \method also achieves 100.0\% on Object and 98.6\% on Goal, supporting the effectiveness of latent future conditioning across diverse manipulation tasks.

\subsection{Evaluation on RoboTwin 2.0}

RoboTwin 2.0 is a bimanual manipulation benchmark with 50 tasks. Following prior work~\citep{bi2026motus,li2026causal,yuan2026fast}, we train a single policy on all tasks using 2,500 demonstrations collected in clean scenes and 25,000 demonstrations collected under heavy domain randomization.

As shown in Table~\ref{tab:robotwin}, \method achieves an average success rate of 91.47\%, improving over $\pi_{0.5}$ by 11.09 percentage points, with consistent gains in both clean (81.12\% to 91.46\%) and randomized (79.64\% to 91.48\%) scenes. Compared with LaWAM, which also models dynamics in latent space, \method achieves a higher average success rate (91.47\% vs.\ 91.22\%) and remains more robust under randomization (91.48\% vs.\ 89.80\%). Per-task results in Appendix Table~\ref{tab:robotwin_per_task} further show large gains over $\pi_{0.5}$ on long-horizon tasks with multiple dependent subgoals, such as Stack Blocks Three (+23.0) and Stack Bowls Three (+33.0), suggesting that anticipating the consequences of latent action plans benefits sequential manipulation.

\begin{table}[h]
\centering
\caption{\textbf{Success rates (\%) on LIBERO-Plus and LIBERO-Pro.} Bold and underlined values indicate the best and second-best results, respectively. $\Delta$ denotes improvement over $\pi_{0.5}$.}
\label{tab:libero_plus_pro}
\resizebox{\linewidth}{!}{%
\begin{tabular}{lccccccccc}
\mytoprule
& \multicolumn{8}{c}{\textbf{LIBERO-Plus}}
& \multicolumn{1}{c}{\textbf{LIBERO-Pro}} \\
\cmidrule(lr){2-9}
\cmidrule(lr){10-10}
\textbf{Method}
& \textbf{Camera}
& \textbf{Robot}
& \textbf{Language}
& \textbf{Light}
& \textbf{Background}
& \textbf{Noise}
& \textbf{Layout}
& \textbf{Avg.}
& \textbf{Avg.} \\
\mymidrule
OpenVLA
& 0.8 & 3.5 & 23.0 & 8.1 & 34.8 & 15.2 & 28.5 & 15.6
& 52.0 \\
Fast-WAM
& 16.4 & 44.5 & 68.9 & 78.2 & 53.7 & 37.7 & 60.7 & 51.5
& 40.2 \\
ImageWAM
& \textbf{80.8} & 50.3 & \textbf{91.4} & \underline{98.1}
& 85.5 & \textbf{93.8} & 80.5 & 83.1
& 48.7 \\
VLA-JEPA
& 63.3 & 67.1 & 85.4 & 95.6 & 93.6 & 66.3 & 85.1 & 79.5
& 48.3 \\
JEPA-WAM
& \underline{79.2} & 59.2 & 68.2 & 93.3 & 94.6 & 83.6 & 76.1 & 79.2
& 41.9 \\
LaWAM
& 37.2 & 71.6 & \underline{87.1} & 93.5 & 94.1 & 69.4 & 83.3 & 76.6
& 52.7 \\
\mymidrule
$\pi_{0.5}$
& 69.4 & \underline{75.3} & 82.6 & 96.7
& \underline{96.8} & \underline{84.3} & \underline{86.2} & \underline{84.5}
& \underline{53.0} \\
\rowcolor{oursgreen}
\textbf{\method (Ours)}
& 75.9 & \textbf{89.3} & 86.1 & \textbf{98.4}
& \textbf{98.1} & 80.6 & \textbf{87.7} & \textbf{88.0}
& \textbf{61.4} \\
\textsl{$\Delta$ Improvement}
& \impv{+6.5} & \impv{+14.0} & \impv{+3.5} & \impv{+1.7}
& \impv{+1.3} & \dropv{-3.7} & \impv{+1.5} & \impv{+3.5}
& \impv{+8.4} \\
\mybottomrule
\end{tabular}%
}
\end{table}

\subsection{Out-of-Distribution Generalization}

We evaluate generalization on LIBERO-Plus, LIBERO-Pro, and LIBERO-Para, directly using the model trained on LIBERO without additional fine-tuning. LIBERO-Plus evaluates robustness across seven perturbation categories, including camera viewpoints, robot initial states, language instructions, lighting, backgrounds, sensor noise, and object layouts. As shown in Table~\ref{tab:libero_plus_pro}, \method achieves an overall success rate of 88.0\%, outperforming all compared methods and exceeding $\pi_{0.5}$ by 3.5 percentage points. It improves performance in six of the seven categories, with the largest gains under robot initial-state perturbations (+14.0) and camera viewpoint changes (+6.5).

On LIBERO-Pro, which measures generalization under changes to objects, spatial configurations, instructions, and environments, \method achieves 61.4\% success, the highest among the compared methods and an 8.4-point improvement over $\pi_{0.5}$. We further assess robustness to instruction paraphrasing on LIBERO-Para, where \method improves success from 71.4\% to 75.0\%. Together, these results indicate that latent future conditioning improves generalization under visual, spatial, and linguistic distribution shifts.

\subsection{Evaluation on Real-World Tasks}

\begin{figure}[t]
  \centering
  \includegraphics[width=\linewidth]{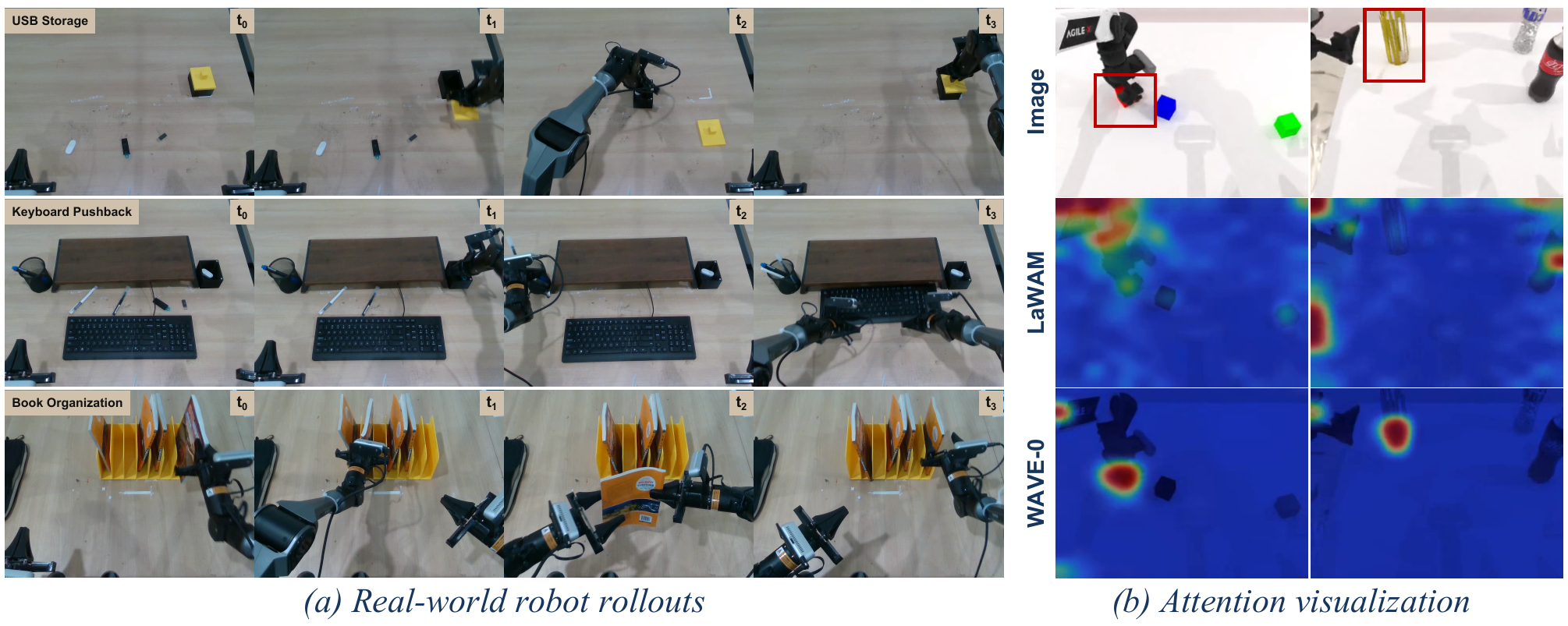}
  \caption{\textbf{Real-world rollouts and attention visualization.} (a) Key frames from \method rollouts on the three real-world tasks. (b) Attention maps on two RoboTwin 2.0 tasks.}
  \label{fig:visualize}
  % \vspace{-1.5em}
\end{figure}

\begin{table}[htbp]
\centering
\footnotesize
% \setlength{\abovecaptionskip}{2pt}
% \setlength{\belowcaptionskip}{4pt}
% \vspace{-0.8em}
% ==================== Left table ====================
\begin{minipage}[t]{0.56\textwidth}
\centering
\caption{\textbf{Real-world success rates (\%).} Each task is evaluated over 20 trials.}
\label{tab:real_robot_results}
\setlength{\tabcolsep}{4pt}
\begin{tabular}{@{}lcccc@{}}
\mytoprule
\textbf{Method} & \textbf{Book} & \textbf{USB} & \textbf{Keyboard} & \textbf{Avg.} \\
\mymidrule
ImageWAM & \underline{50.0} & 25.0 & 45.0 & 40.0 \\
LaWAM & 40.0 & 35.0 & \underline{55.0} & \underline{43.3} \\
\mymidrule
$\pi_{0.5}$ & 40.0 & \underline{45.0} & 30.0 & 38.3 \\
\rowcolor{oursgreen}
\textbf{\method} & \textbf{85.0} & \textbf{65.0} & \textbf{70.0} & \textbf{73.3} \\
\textsl{$\Delta$ Improvement} & \impv{+45.0} & \impv{+20.0} & \impv{+40.0} & \impv{+35.0} \\
\mybottomrule
\end{tabular}
\end{minipage}
\hfill
% ==================== Right table ====================
\begin{minipage}[t]{0.42\textwidth}
\centering
\caption{\textbf{Comparison of training sample efficiency.}}
\label{tab:training_eff}
\setlength{\tabcolsep}{3pt}
\renewcommand{\arraystretch}{1.2}
\begin{tabular}{@{}lcc@{}}
\mytoprule
\textbf{Training stage} & \textbf{LaWAM} & \textbf{\method} \\
\mymidrule
Shared pretraining & 307.20M & 10.24M \\
Policy per task & 12.80M & 7.68M \\
\mymidrule
Total for three tasks & 345.60M & 33.28M \\
\textbf{Amortized per task} & \textbf{115.20M} & \textbf{11.09M} \\
\mybottomrule
\end{tabular}
\end{minipage}
% % ==================== Right table ====================
% \begin{minipage}[t]{0.42\textwidth}
% \centering
% \caption{\textbf{Comparison of training sample efficiency.}}
% \label{tab:training_eff}
% \setlength{\tabcolsep}{4pt}
% \renewcommand{\arraystretch}{1.2}
% \begin{tabular}{@{}lccc@{}}
% \mytoprule
% \textbf{Method} & \textbf{Batch size} & \textbf{Steps} & \textbf{Samples} \\
% \mymidrule
% ImageWAM & 128 & 60k & 7.68M \\
% LaWAM & 128 & 100k & 12.80M \\
% \mymidrule
% $\pi_{0.5}$ & 256 & 30k & 7.68M \\
% \rowcolor{oursgreen}
% \textbf{\method} & 256 & 30k & 7.68M \\
% \mybottomrule
% \end{tabular}
% \end{minipage}
% \vspace{-1.0em}
\end{table}

We evaluate \method on three real-world tasks with a dual-arm AgileX Piper platform, namely Book Organization, USB Storage, and Keyboard Pushback, which require rearranging objects in clutter, precise insertion, and long-horizon bimanual coordination, respectively. For each task, we collect 200 demonstrations, train a separate model for each method on the same data, and evaluate it over 20 trials with diverse initial configurations (Figure~\ref{fig:visualize}(a)). As shown in Table~\ref{tab:real_robot_results}, \method achieves the best performance on all three tasks, reaching an average success rate of 73.3\% compared with 43.3\% for the strongest baseline LaWAM and 40.0\% for ImageWAM. Relative to $\pi_{0.5}$, \method nearly doubles the average success rate from 38.3\% to 73.3\%.

\subsection{Analysis}

\textbf{Training Efficiency.}~
\method learns latent dynamics from action-labeled target demonstrations without large-scale video pretraining. In contrast, LaWAM performs 100k steps of world model pretraining and 200k steps of policy-integration pretraining with a global batch size of 1,024, processing 307.2M examples before benchmark-specific training. On RoboTwin 2.0, the two-stage training of \method requires approximately 670 H100 GPU hours and processes 76.8M examples, compared with 1,280 H100 GPU hours and 102.4M examples for LaWAM's benchmark-specific training alone. Across three real-world tasks, \method processes 33.28M examples, compared with 345.60M for LaWAM, including shared pretraining and task-specific policy training. Amortized per task, these budgets are 11.09M and 115.20M, respectively, with \method achieving higher success rates, as shown in Tables~\ref{tab:training_eff} and~\ref{tab:real_robot_results}. These results support the efficiency of action-grounded latent world modeling without additional large-scale video pretraining.

\textbf{Attention Visualization.}~
Figure~\ref{fig:visualize}(b) compares attention from action queries to current main-view image patches at the final denoising step. In the illustrated examples, \method exhibits more localized attention around the manipulated object and target bottle, whereas LaWAM shows more dispersed responses over the robot arm, image boundaries, and background regions.

\subsection{Ablation Studies}

\subsubsection{Effects of World Model Integration Strategy}

We investigate how the learned world model should be integrated into the policy by comparing $\pi_{0.5}$, $\pi_{0.5}$ + WM, and \method on RoboTwin 2.0. As illustrated in Figure~\ref{fig:comparison}, $\pi_{0.5}$ + WM uses the world model to predict future representations from generated actions and provides an auxiliary training loss against the observed future. In contrast, \method explicitly conditions action generation on predicted futures during both training and inference. Both variants use the same frozen world model trained in Stage~1.

As shown in Table~\ref{tab:robotwin}, auxiliary supervision improves average success from 80.38\% to 89.51\%, supporting the value of the learned latent dynamics for policy learning. \method further increases success to 91.47\%, outperforming $\pi_{0.5}$ + WM by 1.98 and 1.94 percentage points in clean and randomized settings, respectively. These results indicate that while world model supervision already yields substantial gains, explicitly conditioning action generation on predicted futures provides consistent additional benefits. More broadly, they support the central premise of WAM that anticipating the consequences of actions improves action generation, and show that this benefit can be realized in a compact latent space without video generation.

\subsubsection{Effects of Training Objectives}

\begin{wrapfigure}[18]{r}{0.5\textwidth}
\centering
\includegraphics[width=\linewidth]{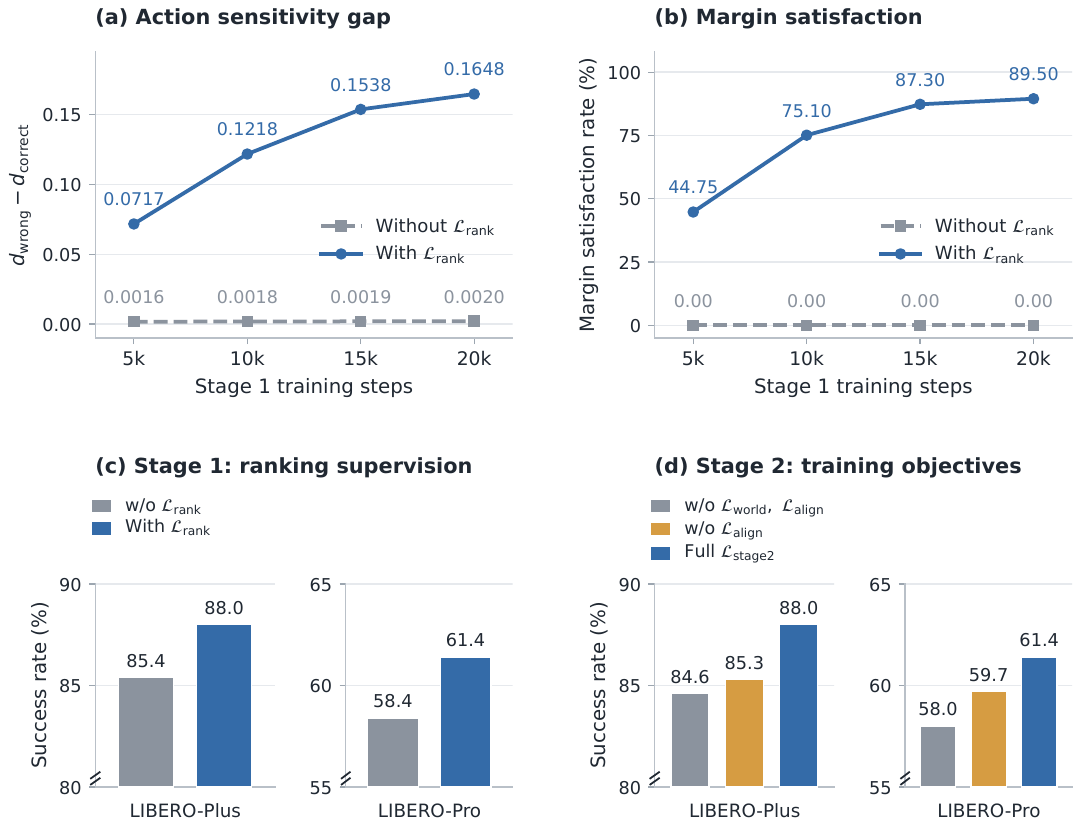}
\caption{\textbf{Effects of training objectives.}}
\label{fig:loss_ablations}
\end{wrapfigure}

\textbf{Counterfactual ranking in Stage~1.}
We remove $\mathcal L_{\mathrm{rank}}$ while retaining $\mathcal L_{\mathrm{future}}$ and $\mathcal L_{\mathrm{delta}}$, and evaluate both variants on the same 2,048 samples from 256 episodes. Holding observations and robot states fixed, we measure the cosine prediction errors under demonstrated and mismatched actions against the same ground-truth future, denoted by $d_{\mathrm{correct}}$ and $d_{\mathrm{wrong}}$. We report the mean gap $g=d_{\mathrm{wrong}}-d_{\mathrm{correct}}$ over valid samples and the percentage satisfying $g\geq 0.1$, termed the action sensitivity gap and margin satisfaction rate, respectively. As shown in Figure~\ref{fig:loss_ablations}(a,b), ranking supervision increases the gap to 0.1648 and margin satisfaction to 89.5\% at 20k steps, compared with 0.0020 and 0\% without it. It also improves success by 2.6 and 3.0 percentage points on LIBERO-Plus and LIBERO-Pro, respectively, supporting the value of action-dependent dynamics for generalization, as shown in Figure~\ref{fig:loss_ablations}(c).

\textbf{Auxiliary objectives in Stage~2.}
Starting from $\mathcal L_{\mathrm{flow}}$ alone, adding $\mathcal L_{\mathrm{world}}$ improves success by 0.7 and 1.7 percentage points on LIBERO-Plus and LIBERO-Pro, respectively. Adding $\mathcal L_{\mathrm{align}}$ yields further gains of 2.7 and 1.7 points, reaching 88.0\% and 61.4\%, as shown in Figure~\ref{fig:loss_ablations}(d). These results support the complementary benefits of supervising latent plans through predicted visual consequences and alignment with learned action representations.

\section{Conclusion}
\label{sec:conclusion}

We presented \method, a framework that bridges VLA policies and world action models through a lightweight latent world model. Instead of generating future frames or using predicted futures only as auxiliary training targets, \method translates VLM-generated latent action plans into predicted future visual representations that explicitly condition the action expert during both training and inference. The world model is grounded in ground-truth robot actions and trained with counterfactual ranking supervision, which encourages action-dependent dynamics and improves policy generalization. Built on $\pi_{0.5}$, \method substantially improves performance on simulation and real-world benchmarks without large-scale world model pretraining, suggesting that grounding latent dynamics in executable actions offers an efficient path toward world-model-augmented robot policies.

\clearpage
\nocite{*}
\bibliographystyle{plainnat}
\bibliography{main}

\appendix
\section{Appendix}

\subsection{Training and Evaluation Details on Simulated Benchmarks}
\label{appendix:training_evaluation}
\textbf{LIBERO.}~
All models are trained on 8 NVIDIA RTX PRO 6000 Blackwell GPUs with 96 GB memory per GPU, using Accelerate DDP, bf16 mixed precision, and the AdamW optimizer. We use a global batch size of 512, a peak learning rate of $1\times10^{-4}$, weight decay of $1\times10^{-2}$, and gradient clipping of 1.0. The learning rate is linearly warmed up for 1,000 steps followed by cosine decay. Training consists of two stages: the latent world model is pretrained for 20k steps, followed by 10k steps of action-policy training with the pretrained world model frozen. Stage 1 takes approximately 10 hours, while Stage 2 takes approximately 13 hours. For LIBERO, we use two camera views, a 7-dimensional action space, and an 8-dimensional proprioceptive state. The policy predicts 32-step action chunks and replans every 10 steps.

\begin{table}[!htbp]
\centering
\caption{\textbf{Training details of WAVE-0 on LIBERO.}}
\label{tab:libero_training_details}
\begin{tabular}{lcc}
\mytoprule
\textbf{Configuration} & \textbf{Stage 1: World Model} & \textbf{Stage 2: Action Policy} \\
\mymidrule
Number of GPUs & 8 & 8 \\
Precision & bf16 & bf16\\
Optimizer & AdamW & AdamW\\
Global batch size & 512 & 512 \\
Peak learning rate & $1\times10^{-4}$ & $1\times10^{-4}$ \\
Final learning rate & $2.5\times10^{-6}$ & $1\times10^{-6}$ \\
Weight decay & $1\times10^{-2}$ & $1\times10^{-2}$ \\
Warmup steps & 1,000 & 1,000 \\
Gradient clipping & 1.0 & 1.0 \\
Training steps & 20,000 & 10,000 \\
Action chunk size $H$& 32 & 32 \\
World model horizon $h$ & 9 & 9 \\
Latent action queries & -- & 10 \\
World model & Trainable & Frozen \\
\mybottomrule
\end{tabular}
\end{table}

\textbf{RoboTwin 2.0.}~
We train \method on 8 NVIDIA H100 GPUs using Accelerate DDP and bf16 mixed precision, with a global batch size of 512. Stage~1 takes approximately 6 hours for 50k steps, while Stage~2 takes approximately 78 hours for 100k steps, with the world model frozen. Observations include three camera views and a 14-dimensional proprioceptive state, with a 14-dimensional action space. Actions and proprioceptive states are normalized using dataset quantiles. The policy predicts and executes 32-step action chunks.

In Stage~2, $\lambda_{\mathrm{WAM}}$ warms up to 1.0 over 1,000 steps. We keep $\lambda_{\mathrm{align}}=1.0$ for the first 5k steps, decay it to 0.1 using a cosine schedule over the next 15k steps, and hold it constant thereafter. We evaluate all 50 tasks using 100 episodes per task under each of the clean and randomized settings. Success rates are averaged over tasks, and the overall score is the mean across both settings. Table~\ref{tab:robotwin_training_details} summarizes the training configuration.

\begin{table}[!htbp]
\centering
\caption{\textbf{Training details of \method on RoboTwin 2.0.}}
\label{tab:robotwin_training_details}
\begin{tabular}{lcc}
\mytoprule
\textbf{Configuration}
& \textbf{Stage 1: World Model}
& \textbf{Stage 2: Action Policy} \\
\mymidrule
Number of GPUs & 8 & 8 \\
Precision & bf16 & bf16 \\
Optimizer & AdamW & AdamW \\
Global batch size & 512 & 512 \\
Peak learning rate & $1\times10^{-4}$ & $1\times10^{-4}$ \\
Final learning rate & $2.5\times10^{-6}$ & $1\times10^{-6}$ \\
Weight decay & $1\times10^{-2}$ & $1\times10^{-2}$ \\
Warmup steps & 1,000 & 1,000 \\
Gradient clipping & 1.0 & 1.0 \\
Training steps & 50,000 & 100,000 \\
Action chunk size $H$ & 32 & 32 \\
World model horizon $h$ & 31 & 31 \\
Latent action queries & -- & 32 \\
World model & Trainable & Frozen \\
\mybottomrule
\end{tabular}
\end{table}

\subsection{Experimental Details on Real-World Tasks}
\label{appendix:real_robot_training_details}

\begin{figure}[t]
  \centering
  \includegraphics[width=\linewidth]{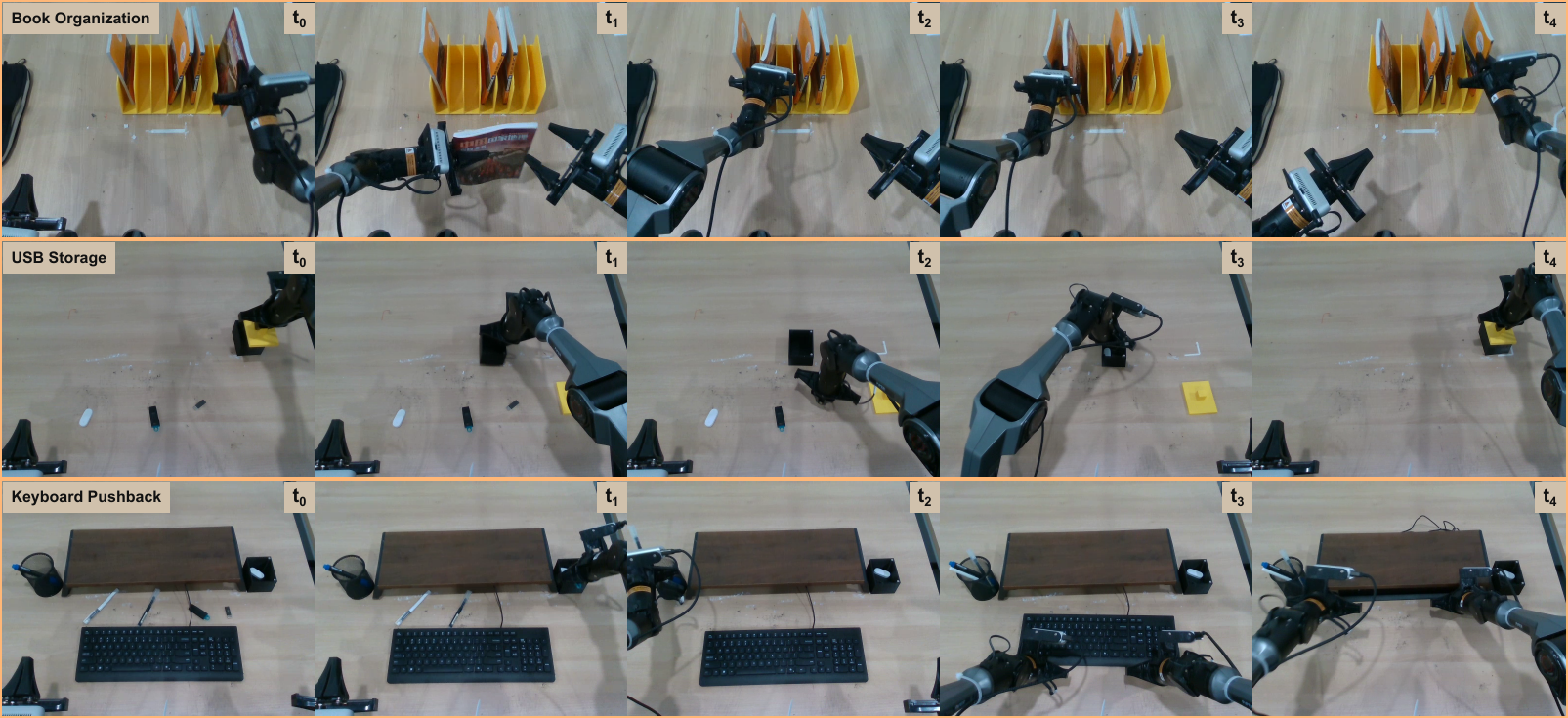}
  \caption{\textbf{Representative rollouts of \method on three real-world manipulation tasks.}}
  \label{fig:realrobot}
\end{figure}

\textbf{Real-robot experiments.}~
We evaluate $\pi_{0.5}$, \method, LaWAM, and ImageWAM on a dual-arm Piper robot across three real-world manipulation tasks: Book Organization, USB Storage, and Keyboard Pushback. Book Organization requires placing a book into the designated book organizer, USB Storage requires transferring a USB drive from the tabletop into a designated storage container, and Keyboard Pushback requires pushing the keyboard back into the drawer. Figure~\ref{fig:realrobot} shows representative rollouts of \method. The complete real-robot dataset contains 600 demonstrations, with 200 demonstrations per task, comprising 310,099 frames recorded at a nominal frequency of 15~Hz. Each observation includes RGB images from one fixed head camera and two wrist cameras, together with a 14-dimensional proprioceptive state. Actions are represented as 14-dimensional absolute joint commands. Each task-specific policy is trained on its corresponding 200 demonstrations. Training is performed on eight NVIDIA A800 GPUs with 80~GB of memory per GPU, using distributed data parallelism, BF16 mixed precision, and the AdamW optimizer. During closed-loop evaluation, $\pi_{0.5}$ and \method predict 32-step action chunks, whereas LaWAM and ImageWAM output 18-step and 16-step action chunks, respectively. $\pi_{0.5}$ and \method execute 10 actions per policy inference, while LaWAM and ImageWAM execute 8.

\FloatBarrier
\paragraph{$\pi_{0.5}$.}
For each real-robot task, we independently fine-tune $\pi_{0.5}$ on the corresponding 200 demonstrations for 30k training steps, using a global batch size of 256. We maintain an exponential moving average of the policy parameters with a decay rate of 0.99 and use the resulting EMA checkpoints for real-robot evaluation. The policy predicts 32-step action chunks and executes 10 actions before replanning. The optimization configuration is summarized in Table~\ref{tab:pi05_real_training}.

\begin{table}[!htbp]
\centering
\caption{\textbf{Training details of $\pi_{0.5}$ in the real-world experiments.}}
\label{tab:pi05_real_training}
\begin{tabular}{lc}
\mytoprule
\textbf{Configuration}
& \textbf{$\pi_{0.5}$} \\
\mymidrule
Number of GPUs & 8 \\
Precision & bf16 \\
Optimizer & AdamW \\
Global batch size & 256 \\
Peak learning rate & $2.5\times10^{-5}$ \\
Final learning rate & $2.5\times10^{-6}$ \\
Weight decay & $1\times10^{-10}$ \\
Warmup steps & 1,500 \\
Gradient clipping & 1.0 \\
Training steps & 30,000 \\
Action chunk size $H$ & 32 \\
\mybottomrule
\end{tabular}
\end{table}

\paragraph{\method.}
\method is trained in two stages. In Stage~1, the latent world model is pretrained for 40k steps using all 600 real-robot demonstrations collected from the three manipulation tasks. In Stage~2, the pretrained world model is frozen, and a task-specific action policy is independently trained for 30k steps using the corresponding 200 demonstrations from each task. Both stages use linear learning-rate warmup followed by cosine decay. The world model predicts visual representations 32 steps into the future, while the action policy generates 32-step action chunks using 32 latent action queries. The detailed training configuration is summarized in Table~\ref{tab:wave0_real_training_details}.

\begin{table}[!htbp]
\centering
\caption{\textbf{Training details of \method in the real-world experiments.}}
\label{tab:wave0_real_training_details}
\begin{tabular}{lcc}
\mytoprule
\textbf{Configuration}
& \textbf{Stage 1: World Model}
& \textbf{Stage 2: Action Policy} \\
\mymidrule
Number of GPUs & 8 & 8 \\
Precision & bf16 & bf16 \\
Optimizer & AdamW & AdamW \\
Global batch size & 256 & 256 \\
Peak learning rate & $1\times10^{-4}$ & $5\times10^{-5}$ \\
Final learning rate & $2.5\times10^{-6}$ & $1\times10^{-6}$ \\
Weight decay & $1\times10^{-2}$ & $1\times10^{-2}$ \\
Warmup steps & 2,000 & 1,500 \\
Gradient clipping & 1.0 & 1.0 \\
Training steps & 40,000 & 30,000 \\
Action chunk size $H$ & 32 & 32 \\
World model horizon $h$ & 32 & 32 \\
Latent action queries & -- & 32 \\
World model & Trainable & Frozen \\
\mybottomrule
\end{tabular}
\end{table}

\paragraph{LaWAM.}
For each real-robot task, LaWAM is independently trained on the corresponding 200 demonstrations for 100k optimization steps. Training is performed on eight GPUs with a per-GPU batch size of 16, resulting in a global batch size of 128. The learning rate is linearly warmed up for 500 steps and subsequently decayed using a cosine schedule. LaWAM uses three RGB camera views as input and does not condition its flow-matching action head on the robot state. Its internal flow horizon is 50, from which the first 18 valid actions are returned during inference. The main training parameters are summarized in Table~\ref{tab:lawam_real_training_details}.

\paragraph{ImageWAM.}
For each real-robot task, ImageWAM is independently trained on the corresponding 200 demonstrations for 60k optimization steps. Training is performed on eight GPUs with a per-GPU batch size of 16, resulting in a global batch size of 128.  We train the action prediction module using BF16 mixed precision and the AdamW optimizer. The learning rate is linearly warmed up for 3,000 steps, corresponding to 5\% of the total training budget, and then decayed using a cosine schedule. ImageWAM jointly learns visual dynamics and action prediction from the three camera views and produces a 16-step absolute joint-action chunk during inference. The main training parameters are summarized in Table~\ref{tab:imagewam_real_training_details}.

\begin{table}[!htbp]
\centering

\begin{minipage}[t]{0.48\linewidth}
\centering
% \vspace{0pt}
\caption{\textbf{Training details of LaWAM in the real-world experiments.}}
\label{tab:lawam_real_training_details}
\small
\setlength{\tabcolsep}{4pt}
\begin{tabular}{@{}lc@{}}
\mytoprule
\textbf{Configuration} & \textbf{LaWAM} \\
\mymidrule
Number of GPUs & 8 \\
Precision & bf16 \\
Optimizer & AdamW \\
Per-GPU batch size & 16 \\
Global batch size & 128 \\
Peak learning rate & $1\times10^{-4}$ \\
Final learning rate & $5\times10^{-7}$ \\
Weight decay & $1\times10^{-8}$ \\
Warmup steps & 500 \\
Gradient clipping & 1.0 \\
Training steps & 100,000 \\
Internal flow horizon & 50 \\
Output action horizon $H$ & 18 \\
\mybottomrule
\end{tabular}
\end{minipage}
\hfill
\begin{minipage}[t]{0.48\linewidth}
\centering
% \vspace{0pt}
\caption{\textbf{Training details of ImageWAM in the real-world experiments.}}
\label{tab:imagewam_real_training_details}
\small
\setlength{\tabcolsep}{4pt}
\begin{tabular}{@{}lc@{}}
\mytoprule
\textbf{Configuration} & \textbf{ImageWAM} \\
\mymidrule
Number of GPUs & 8 \\
Precision & bf16 \\
Optimizer & AdamW \\
Per-GPU batch size & 16 \\
Global batch size & 128 \\
Peak learning rate & $1\times10^{-4}$ \\
Final learning rate & $1\times10^{-6}$ \\
Weight decay & $1\times10^{-2}$ \\
Warmup steps & 3,000 \\
Gradient clipping & 1.0 \\
Training steps & 60,000 \\
Action chunk size $H$ & 16 \\
\mybottomrule
\end{tabular}
\end{minipage}
\end{table}

\textbf{Inference efficiency.}~
Our lightweight latent world model adds only modest inference overhead to $\pi_{0.5}$. As shown in Table~\ref{tab:real_robot_latency}, \method increases the mean server-side latency from 287.1~ms to 305.9~ms, an additional 18.8~ms, while enabling explicit future conditioning during action generation. Compared with ImageWAM, an efficient visual generation-based WAM requiring 582.6~ms per request, \method reduces latency by 47.5\%.

LaWAM achieves a lower latency of 223.9~ms with a smaller model of 2.3B parameters, compared with 4.3B for \method. It also returns 18 actions per inference request, whereas \method predicts 32. These differences in model size and inference configuration should be considered when interpreting the latency comparison.

\begin{table}[!htbp]
\centering
\caption{\textbf{Mean server-side inference latency in the real-world experiments.}}
\label{tab:real_robot_latency}
\begin{tabular}{lcc}
\mytoprule
\textbf{Model} & \textbf{Checkpoint} & \textbf{Mean server latency} \\
\mymidrule
$\pi_{0.5}$    & 30k  & 287.1~ms \\
LaWAM    & 100k & \textbf{223.9~ms} \\
ImageWAM & 60k  & 582.6~ms \\
\method   & 30k  & 305.9~ms \\
\mybottomrule
\end{tabular}
\end{table}

\subsection{LIBERO-Pro Detailed Results}
\begin{table}[htbp]
\caption{\textbf{Results on LIBERO-Pro across different perturbation categories.}}
\label{tab:libero-pro}
\begin{center}

\resizebox{\textwidth}{!}{
\begin{tabular}{lc|ccccc|ccccc|ccccc|ccccc}
\mytoprule

\multirow{2}{*}{\bf Method} &
\multirow{2}{*}{\bf Avg.} &
\multicolumn{5}{c|}{\bf Spatial} &
\multicolumn{5}{c|}{\bf Object} &
\multicolumn{5}{c|}{\bf Goal} &
\multicolumn{5}{c}{\bf Long} \\

& &
{\bf Obj.} & {\bf Pos.} & {\bf Sem.} & {\bf Task} & {\bf Env.} &
{\bf Obj.} & {\bf Pos.} & {\bf Sem.} & {\bf Task} & {\bf Env.} &
{\bf Obj.} & {\bf Pos.} & {\bf Sem.} & {\bf Task} & {\bf Env.} &
{\bf Obj.} & {\bf Pos.} & {\bf Sem.} & {\bf Task} & {\bf Env.} \\

\mymidrule

OpenVLA
& 52.0
& 97.0 & 0.0 & \underline{97.0} & 0.0 & \textbf{89.0}
& \textbf{98.0} & 0.0 & 98.0 & 0.0 & 0.0
& \underline{96.0} & 0.0 & 98.0 & 0.0 & \textbf{98.0}
& \underline{81.0} & 0.0 & 96.0 & 0.0 & \textbf{85.0} \\

$\pi_0$
& 44.0
& 95.0 & 0.0 & \underline{97.0} & 0.0 & 60.0
& \underline{94.0} & 0.0 & 90.0 & 0.0 & 29.0
& 94.0 & 0.0 & 93.0 & 0.0 & 39.0
& 79.0 & 0.0 & 82.0 & 0.0 & 27.0 \\

$\pi_{0.5}$
& \underline{53.0}
& 97.0 & 20.0 & \underline{97.0} & 0.01 & 46.0
& \textbf{98.0} & 0.2 & 96.0 & 0.0 & 0.7
& \textbf{97.0} & \textbf{38.0} & 97.0 & 0.0 & 46.0
& \textbf{92.0} & 0.1 & 93.0 & 0.0 & 46.0 \\

Fast-WAM
& 40.2
& 87.0 & 0.0 & 58.0 & 32.2 & 36.2
& 93.2 & 0.8 & 98.6 & 9.2 & 35.4
& 56.6 & 0.2 & 92.2 & 8.6 & 7.6
& 52.4 & 0.0 & 79.4 & 9.6 & 45.8 \\

ImageWAM
& 48.7
& 95.6 & 18.0 & 95.2 & 49.6 & 56.6
& 91.0 & 0.0 & \underline{99.6} & 6.6 & 40.6
& 68.2 & 3.4 & \underline{98.2} & 10.0 & 31.4
& 53.6 & 0.0 & 96.2 & 10.0 & 50.2 \\

LaWAM
& 52.7
& \underline{97.8} & \underline{28.8} & 93.2 & \textbf{52.6} & 43.4
& 91.8 & \underline{10.6} & \underline{99.6} & \underline{10.0} & \underline{61.0}
& 68.4 & 9.0 & \textbf{98.4} & \textbf{28.4} & 25.6
& 62.8 & \underline{1.2} & \underline{96.6} & \underline{23.0} & 51.2 \\

VLA-JEPA
& 48.3
& 96.6 & 13.4 & 93.0 & \underline{51.4} & 22.6
& 90.2 & 0.00 & \textbf{99.8} & \underline{10.0} & 49.8
& 77.4 & 2.20 & \underline{98.2} & 12.0 & 35.4
& 53.2 & 0.2 & 91.0 & 9.2 & 59.6 \\

JEPA-WAM
& 41.9
& 85.4 & 0.2 & 69.0 & 51.6 & 35.0
& 92.8 & 0.0 & 99.0 & 10.0 & 14.0
& 57.8 & 0.0 & 72.6 & 11.2 & 46.6
& 36.8 & 0.0 & 85.4 & 9.4 & 61.6 \\

\mymidrule
\rowcolor{oursgreen}
\textbf{WAVE-0 (Ours)}
& \textbf{61.4}
& \textbf{98.0} & \textbf{54.6} & \textbf{97.4} & 48.0 & \underline{64.2}
& 83.0 & \textbf{54.2} & \textbf{99.8} & \textbf{13.2} & \textbf{70.4}
& 85.6 & \underline{23.4} & 97.8 & \underline{21.4} & \underline{47.0}
& 70.2 & \textbf{11.4} & \textbf{97.2} & \textbf{27.6} & \underline{63.4} \\

\mybottomrule
\end{tabular}
}

\end{center}
\end{table}

\subsection{RoboTwin Detailed Results}

\begin{table}[p]
\centering
\small
\setlength{\tabcolsep}{3pt}
\renewcommand{\arraystretch}{1.05}
\caption{\textbf{Per-task success rates (\%) on RoboTwin 2.0.}}
\label{tab:robotwin_per_task}
\begin{adjustbox}{
  max totalsize={\textwidth}{0.88\textheight}
}
\begin{tabular}{lcc|cc|cc|cc|cc|cc}
\mytoprule
\multirow{2}{*}{\textbf{Task}} & \multicolumn{2}{c}{\textbf{\method~(Ours)}} & \multicolumn{2}{c}{$\pi_{0.5}$} & \multicolumn{2}{c}{Motus} & \multicolumn{2}{c}{Fast-WAM} & \multicolumn{2}{c}{LingBot-VA} & \multicolumn{2}{c}{LaWAM} \\
\cmidrule(lr){2-3}\cmidrule(lr){4-5}\cmidrule(lr){6-7}\cmidrule(lr){8-9}\cmidrule(lr){10-11}\cmidrule(lr){12-13}
 & Clean & Rand. & Clean & Rand. & Clean & Rand. & Clean & Rand. & Clean & Rand. & Clean & Rand. \\
\mymidrule
Adjust Bottle & \textbf{100} & \textbf{100} & \textbf{100} & 99 & 89 & 93 & \textbf{100} & \textbf{100} & 90 & 94 & \textbf{100} & \textbf{100} \\
Beat Block Hammer & \textbf{100} & \textbf{99} & 92 & 94 & 95 & 88 & 99 & 97 & 96 & 98 & 90 & 93 \\
Blocks Ranking RGB & \textbf{100} & 99 & 92 & 90 & 99 & 97 & \textbf{100} & \textbf{100} & 99 & 98 & 97 & \textbf{100} \\
Blocks Ranking Size & 90 & 87 & 78 & 74 & 75 & 63 & \textbf{94} & \textbf{98} & \textbf{94} & 96 & 93 & 89 \\
Click Alarmclock & \textbf{100} & 99 & \textbf{100} & 99 & \textbf{100} & \textbf{100} & \textbf{100} & \textbf{100} & 99 & \textbf{100} & \textbf{100} & \textbf{100} \\
Click Bell & \textbf{100} & \textbf{100} & \textbf{100} & 98 & \textbf{100} & \textbf{100} & \textbf{100} & \textbf{100} & \textbf{100} & \textbf{100} & \textbf{100} & \textbf{100} \\
Dump Bin Bigbin & 95 & \textbf{98} & 86 & 91 & 95 & 91 & \textbf{97} & 96 & 89 & 96 & \textbf{97} & 95 \\
Grab Roller & \textbf{100} & \textbf{100} & \textbf{100} & \textbf{100} & \textbf{100} & \textbf{100} & \textbf{100} & \textbf{100} & \textbf{100} & \textbf{100} & \textbf{100} & \textbf{100} \\
Handover Block & 97 & \textbf{91} & 59 & 45 & 86 & 73 & 95 & 81 & \textbf{99} & 78 & 96 & 87 \\
Handover Mic & 97 & 99 & 90 & 88 & 78 & 63 & \textbf{99} & \textbf{100} & 94 & 96 & 93 & 98 \\
Hanging Mug & 47 & 49 & 40 & 49 & 38 & 38 & \textbf{58} & \textbf{62} & 40 & 28 & 51 & 43 \\
Lift Pot & \textbf{100} & \textbf{100} & 93 & 88 & 96 & 99 & \textbf{100} & \textbf{100} & \textbf{100} & 99 & \textbf{100} & 99 \\
Move Can Pot & 91 & 94 & 72 & 72 & 34 & 74 & 90 & 88 & 94 & \textbf{97} & \textbf{98} & 93 \\
Move Pillbottle Pad & 98 & 97 & 74 & 80 & 93 & 96 & \textbf{100} & \textbf{99} & 99 & \textbf{99} & 97 & 90 \\
Move Playingcard Away & \textbf{100} & 99 & 97 & 93 & \textbf{100} & 96 & \textbf{100} & \textbf{100} & \textbf{100} & 99 & \textbf{100} & \textbf{100} \\
Move Stapler Pad & 55 & 52 & 42 & 27 & 83 & 85 & 77 & 64 & 91 & 79 & \textbf{94} & \textbf{87} \\
Open Laptop & 97 & 99 & 99 & 98 & 95 & 91 & 98 & \textbf{100} & 92 & 94 & \textbf{100} & \textbf{100} \\
Open Microwave & 58 & 74 & 28 & 14 & \textbf{95} & \textbf{91} & 62 & 45 & 82 & 86 & 41 & 43 \\
Pick Diverse Bottles & 78 & 81 & 82 & 78 & 90 & \textbf{91} & 80 & 85 & 89 & 82 & \textbf{91} & 88 \\
Pick Dual Bottles & 94 & 96 & 97 & 94 & 96 & 90 & \textbf{100} & 96 & \textbf{100} & \textbf{99} & \textbf{100} & 95 \\
Place A2B Left & 91 & \textbf{94} & 93 & 90 & 88 & 79 & 95 & 93 & 97 & 93 & \textbf{98} & 91 \\
Place A2B Right & \textbf{99} & 95 & 97 & 92 & 91 & 87 & 93 & \textbf{99} & 97 & 95 & 89 & 94 \\
Place Bread Basket & 87 & 91 & 76 & 80 & 91 & 94 & 91 & 93 & \textbf{97} & \textbf{95} & 92 & 85 \\
Place Bread Skillet & 90 & 85 & 80 & 73 & 86 & 83 & 90 & \textbf{93} & \textbf{95} & 90 & 90 & 83 \\
Place Burger Fries & \textbf{98} & 94 & 90 & 90 & \textbf{98} & 98 & 96 & \textbf{99} & 97 & 95 & 93 & 96 \\
Place Can Basket & 84 & 76 & 56 & 58 & 81 & 76 & 71 & 69 & 81 & \textbf{84} & \textbf{92} & 65 \\
Place Cans Plasticbox & 97 & \textbf{99} & 79 & 76 & 98 & 94 & 99 & 96 & \textbf{100} & \textbf{99} & \textbf{100} & 95 \\
Place Container Plate & 97 & 99 & 93 & 96 & 98 & 99 & 96 & \textbf{100} & 99 & 97 & \textbf{100} & \textbf{100} \\
Place Dual Shoes & 94 & \textbf{97} & 87 & 80 & 93 & 87 & 94 & 88 & 94 & 89 & \textbf{98} & 94 \\
Place Empty Cup & \textbf{100} & \textbf{100} & 98 & 96 & 99 & 98 & \textbf{100} & \textbf{100} & \textbf{100} & \textbf{100} & 99 & \textbf{100} \\
Place Fan & 91 & 92 & 69 & 59 & 91 & 87 & 96 & \textbf{96} & \textbf{99} & 93 & 92 & 93 \\
Place Mouse Pad & 90 & 85 & 71 & 73 & 66 & 68 & 83 & 89 & \textbf{93} & \textbf{96} & 91 & 84 \\
Place Object Basket & 91 & 89 & 77 & 70 & 81 & 87 & 89 & 88 & 91 & 88 & \textbf{92} & \textbf{90} \\
Place Object Scale & 95 & 95 & 81 & 78 & 88 & 85 & 90 & \textbf{97} & \textbf{96} & 95 & 95 & 88 \\
Place Object Stand & 93 & 95 & 81 & 83 & 98 & \textbf{97} & 90 & 94 & \textbf{99} & 96 & 92 & 93 \\
Place Phone Stand & \textbf{97} & 97 & 92 & 90 & 87 & 86 & \textbf{97} & \textbf{99} & \textbf{97} & 97 & 93 & 94 \\
Place Shoe & 99 & \textbf{100} & 97 & 98 & 99 & 97 & 96 & 99 & 98 & 98 & \textbf{100} & \textbf{100} \\
Press Stapler & 93 & 96 & \textbf{98} & \textbf{99} & 93 & 98 & 90 & 97 & 85 & 82 & \textbf{98} & 97 \\
Put Bottles Dustbin & 90 & 85 & 72 & 61 & 81 & 79 & \textbf{95} & 90 & 87 & 91 & 94 & \textbf{92} \\
Put Object Cabinet & 86 & 83 & 77 & 74 & 88 & 71 & \textbf{94} & \textbf{89} & 85 & 87 & 90 & 82 \\
Rotate QRcode & 93 & \textbf{95} & 80 & 85 & 89 & 73 & 93 & 89 & \textbf{96} & 91 & 94 & 89 \\
Scan Object & 95 & \textbf{95} & 91 & 88 & 67 & 66 & 89 & 92 & \textbf{96} & 91 & \textbf{96} & 90 \\
Shake Bottle & \textbf{100} & 99 & 98 & \textbf{100} & \textbf{100} & 97 & \textbf{100} & \textbf{100} & \textbf{100} & 97 & \textbf{100} & \textbf{100} \\
Shake Bottle Horizontally & \textbf{100} & \textbf{100} & 98 & \textbf{100} & \textbf{100} & 98 & \textbf{100} & \textbf{100} & \textbf{100} & 99 & \textbf{100} & \textbf{100} \\
Stack Blocks Three & 97 & 94 & 77 & 68 & 91 & 95 & 95 & 97 & \textbf{99} & \textbf{98} & 90 & 75 \\
Stack Blocks Two & \textbf{100} & \textbf{100} & 93 & 94 & \textbf{100} & 98 & \textbf{100} & \textbf{100} & \textbf{100} & 98 & \textbf{100} & 97 \\
Stack Bowls Three & 82 & 81 & 48 & 49 & 79 & \textbf{87} & 80 & 81 & 86 & 83 & \textbf{90} & 80 \\
Stack Bowls Two & 93 & 94 & 84 & 81 & 98 & 98 & 92 & 98 & 94 & 98 & \textbf{100} & \textbf{99} \\
Stamp Seal & 84 & 79 & 35 & 56 & 93 & 92 & 90 & 94 & \textbf{96} & \textbf{97} & 89 & 88 \\
Turn Switch & 70 & 77 & 67 & 74 & \textbf{84} & \textbf{78} & 61 & 59 & 44 & 45 & 47 & 56 \\
\mymidrule
\textbf{Average} & 91.46 & 91.48 & 81.12 & 79.64 & 88.66 & 87.02 & 91.88 & \textbf{91.78} & \textbf{92.90} & 91.50 & 92.64 & 89.80 \\
\mybottomrule
\end{tabular}
\end{adjustbox}

\end{table}

\end{document}